# Unsupervised Learning of Depth, Optical Flow and Pose with Occlusion from 3D Geometry

Guangming Wang, Chi Zhang, Hesheng Wang, Jingchuan Wang, Yong Wang, and Xinlei Wang

*Abstract*—In autonomous driving, monocular sequences contain lots of information. Monocular depth estimation, camera ego-motion estimation and optical flow estimation in consecutive frames are high-profile concerns recently. By analyzing tasks above, pixels in the middle frame are modeled into three parts: the rigid region, the non-rigid region, and the occluded region. In joint unsupervised training of depth and pose, we can segment the occluded region explicitly. The occlusion information is used in unsupervised learning of depth, pose and optical flow, as the image reconstructed by depth-pose and optical flow will be invalid in occluded regions. A less-than-mean mask is designed to further exclude the mismatched pixels interfered with by motion or illumination change in the training of depth and pose networks. This method is also used to exclude some trivial mismatched pixels in the training of the optical flow network. Maximum normalization is proposed for depth smoothness term to restrain depth degradation in textureless regions. In the occluded region, as depth and camera motion can provide more reliable motion estimation, they can be used to instruct unsupervised learning of optical flow. Our experiments in KITTI dataset demonstrate that the model based on three regions, full and explicit segmentation of the occlusion region, the rigid region, and the non-rigid region with corresponding unsupervised losses can improve performance on three tasks significantly. The source code is available at: https://github.com/guangmingw/DOPlearning.

## I. INTRODUCTION

IN autonomous driving, it's a key issue to get the depth of the scene and its localization to construct the map. Lasers bring more accurate information [1], [2], but at the same time, it increases costs and needs calibration [3]. Using only inexpensive vision sensors can get dense information, which is also closer to the way people perceive information when driving. However, the traditional visual SLAM method relies heavily on artificial design features [4]. It is also not robust enough for changes in the environment, and it is easy to lose features in dynamic environments and fail outdoors.

A method based on deep learning can obtain dense depth information [1], and do not rely on hand-designed features. In addition, the contrast experiment proposed by Tateno et al. (CNN-SLAM) [5] made with LSD-SLAM [6] shows the advantage of the monocular depth estimation based on deep learning which does not require initialization. However, supervised learning methods [7], [8] rely on expensive truth data, which relies on expensive devices and manual labeling [9]. Recently, unsupervised learning methods are developing, which made this problem more and more independent of ground truth [10]-[23]. Some even only need monocular video sequences, which have recently received much attention [14]-[23]. Scene flow of each pixel in the monocular video can be obtained with optical flow and depth, which indicates the motion information in the three-dimensional (3D) environment [24], so the optical flow estimation [25], [26] is also very important in autonomous driving. The depth-pose can only characterize the static information in the image and the ego-motion of the camera, while the optical flow can characterize the dynamics, just to make up for it. However, the ground truth of optical flow is more difficult to obtain, which makes the unsupervised learning of optical flow highly important [21]-[23], [27]-[30]. Moreover, the occlusion problem is an important issue in the estimation of optical flow [29], [30], and the depth-pose can be used to calculate the occlusion mask [19] explicitly. Therefore, the joint assisted learning of depth, pose and optical flow is full of potential, and our article is to study this.

The main contributions of this paper are as follows: (1) According to the motion information of each pixel in the first frame to the second frame, the first frame is divided into three parts including the occluded region, the rigid region, and the non-rigid region. Then, through the 3D geometric relationship of the point cloud obtained by depth-pose networks, the occlusion problem is considered explicitly, and overlap and blank masks are designed to filter out occluded pixels that do not satisfy the pixel matching in the reconstructed image, which is the main part in the preliminary conference version [19]. In this extension version, we also make a theoretical analysis of the effectiveness of the occlusion mask. The occlusion region also does not satisfy the image reconstruction loss of the optical flow and is filtered out in the training process. And the following contributions are all novel and not included in the preliminary conference version [19]. (2) When computing the

*This work was supported in part by the Natural Science Foundation of China under Grant U1613218, 61722309 and U1913204, in part by Beijing Advanced Innovation Center for Intelligent Robots and Systems under Grant 2019IRS01. Corresponding Author: Hesheng Wang.

G. Wang, C. Zhang, H. Wang and J. Wang are with Department of Automation, Insititue of Medical Robotics, Key Laboratory of System Control and Information Processing of Ministry of Education, Key Laboratory of Marine Intelligent Equipment and System of Ministry of Education, Shanghai Jiao Tong University, Shanghai 200240, China. H. Wang is also with Beijing Advanced Innovation Center for Intelligent Robots and Systems, Beijing Institute of Technology, China.

Y. Wang is with Beijing Institute of Control Engineering, Beijing 100080, China.

X. Wang is with DeepBlue Academy of Sciences, Shanghai 200240, China.

image reconstruction loss of depth-pose networks and optical flow network, a less-than-mean mask is added to further filter outliers. (3) An inverse depth maximum normalization method is proposed for the smoothness loss to prevent the problem of decreasing inverse depth values during training. And a comparative analysis of our normalization method and the normalization method proposed by Wang et al. [16] is made. (4) Since the optical flow cannot find matching pixels from two consecutive frames in the occluded area to perform unsupervised training, consistency loss is proposed, which makes optical flow can be supervised by rigid flow calculated from the depth and pose information in the occluded region. The effectiveness of our system and methods is demonstrated by experiments on the KITTI [9] and Cityscape [31] datasets.

The second section introduces the related work. Our system and methods are presented in the third section. The fourth section is about the experimental validation, and finally, the fifth section draws a conclusion.

## II. Related Work

### A. Unsupervised Learning of Depth and Pose

Unsupervised estimation of monocular depth starts from stereo image pairs [10]. After the depth of an image is estimated by the network, its projection onto another frame can be used for image reconstruction, and unsupervised training is conducted by the consistency of left and right images. Godard et al. [11] propose the depth consistency loss of the left and right cameras. Prior loss of smoothness term [10], [11] is considered to alleviate some wrong estimates because of occlusion. Warping loss based on feature rather than direct pixel values between stereo sequences is adopted by Zhan et al. [12] to learn depth and odometry. UnDeepVO [13] uses stereo sequences for training and is tested on monocular sequences so that depth and pose network can recover spatial scale from monocular sequences.

Zhou et al. [14] implement the joint unsupervised learning of depth and ego-motion using only monocular sequences for the first time and obtain competitive results. The learning cost of depth and pose is further reduced by only using monocular sequences for training, making the data sources of these two networks more extensive and cheaper.

Iterative Closest Point (ICP) [32] is added in vid2depth [15] to perform rigid constraints on the depth estimation in 3D to improve the results. The combination of direct methods of visual odometry and pose network is tried by Wang et al. [16] to improve the effect of pose and depth estimation, and a depth normalization strategy is used to improve the effect of depth estimation. A generative adversarial network is used in GANVO [17] to judge the quality of reconstructed images, so as to train the depth and pose networks. A self-adjusting threshold is adopted in DeepMatchVO [18] to block the image area from participating in the computation of projection error and epipolar geometry is introduced to add constraints.

Zhou et al. [14] use warping between monocular adjacent frames, and output an explainability mask when executing pose estimation, to remove the inconsistent regions between adjacent frames, which contain the information of motion and occlusion. However, their later experiments find that the results are better without this learned explainability mask[1]. After that, Ranjan et al. [23] add a new motion segmentation network to estimate motion segmentation mask combined with the training of optical flow to achieve better results. However, the motion segmentation mask of CC [23] is to exclude the interference of motion, and CC [23] lacks occlusion consideration, which is a key problem in the unsupervised process of optical flow [21]-[23], [27]-[30], and is also an important problem in the training of depth-pose networks through the analysis in our paper. Therefore, we propose a parameterless calculation method for the explicit occlusion mask to solve the occlusion problem. The method is based on 3D geometric relations in the process of image reconstruction. Our proposed occlusion mask excludes the mismatched pixels in the occlusion area and improves the performance of depth-pose networks. Considering the difference magnitude of the pixel matching directly, less-than-mean mask is also proposed in this paper to further exclude the dynamic area and illumination change area compared to CC [23]. We also found the disadvantages of the mean normalization strategy [16] in the depth smoothness term, and a novel maximum normalization strategy is proposed in this paper accordingly.

### B. Unsupervised Learning of Depth, Pose and Optical Flow

With monocular sequence alone being used to estimate depth and pose, combined studies begin to appear. Edge, normal vector and depth are estimated simultaneously in LEGO [20]. Joint learning of depth, pose and optical flow is put forward in GeoNet [21] for the first time, but it only uses optical flow network to estimate residual part after depth and pose training. This method doesn't consider occlusion in the training of depth and pose, and still adopt the occlusion processing method of forward-backward consistency in optical flow training, which needs to adjust the parameters according to the accuracy of optical flow estimation. The two-stage learning of optical flow is similar to the work of Alletto et al. [27], which also divides the learning of optical flow into two parts and proposes a projection self-supervised estimation method of optical flow. Joint training of depth, pose and optical flow is implemented in DF-Net [22], adding the cross loss in the forward-backward consistency region. An image is divided into dynamic area and static area by the motion segmentation mask estimated by a motion segmentation network in CC [23], and the static area is used for the depth-pose training and the dynamic area is for the optical flow training. And all networks are trained by the competitive collaboration [23] of the two parts—the depth-pose and the optical flow. However, the occlusion area does not satisfy the image reconstruction loss in the training of depth-pose or optical flow, so that regardless of which area the occlusion area is divided into, it will bring errors and mislead the training of networks. At the same time, complicated iterative training, in turn, is needed in CC [23]. Therefore, our paper focus on the occlusion problem.

---

[1] https://github.com/tinghuiz/SfMLearner

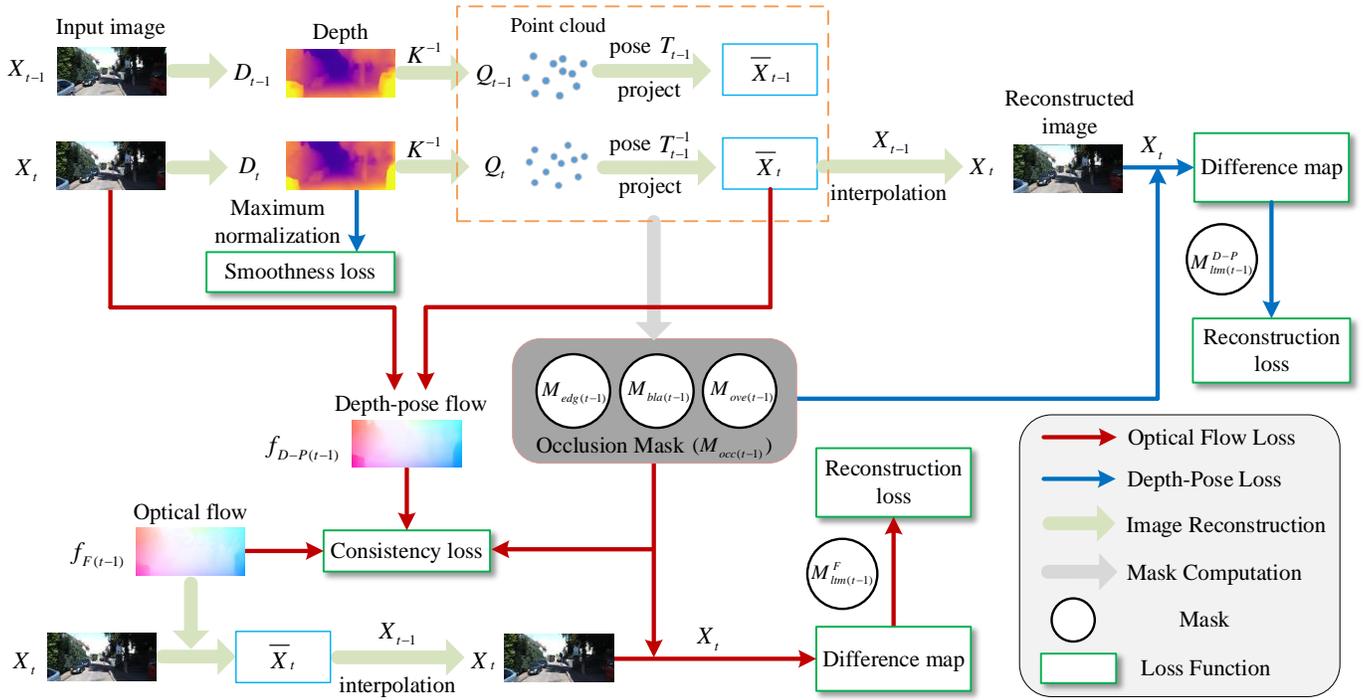

Fig. 1. The overview of our proposed joint unsupervised training pipeline of depth, pose, and optical flow. The image reconstruction loss and the depth smoothness loss are used in the training of the depth-pose networks. The image reconstruction loss, the consistency loss and the optical flow smoothness loss are used in the training of the optical flow network. The depth-pose flow and the occlusion mask in the depth-pose phase are used in the unsupervised training process of the optical flow network.

One important difference between the previous works [21]-[23] and ours is that explicit and parameterless occlusion computation is added by us in the joint learning process of the three to solve the occlusion problem, and its validity is analyzed in theory. Compared with CC [23], we not only consider the important occlusion problem, but also make the model simpler and easier to train with the calculated occlusion mask and less-than-mean mask. What's more, we extend the calculated occlusion mask to the unsupervised learning of optical flow and propose a new consistency loss where depth-pose flow supervises optical flow in the occlusion area for a better flow estimation, which is unique to us because of our explicitly occlusion consideration.

### III. JOINT UNSUPERVISED TRAINING OF DEPTH-POSE AND OPTICAL FLOW NETWORKS

Our system depends on three networks to estimate monocular depth and to obtain the ego-motion and optical flow between two consecutive frames with only monocular video sequences. The three networks include depth network $N_{depth}$, pose network $N_{pose}$ and optical flow network $N_{flow}$. The input of $N_{depth}$ is a single image $X_t$ and output is depth value for each pixel. The inputs of $N_{pose}$ are three consecutive frames $(X_{t-1}, X_t, X_{t+1})$ and outputs are poses between the intermediate frame and the adjacent two frames. The inputs of $N_{flow}$ are three consecutive frames $(X_{t-1}, X_t, X_{t+1})$ and outputs are optical flows from the intermediate frame to the adjacent two frames.

By analyzing the process of the image reconstruction with depth and pose, and the process of the image reconstruction with optical flow, the image is divided into three parts including the rigid region, the non-rigid region, and the occluded region. Corresponding loss functions are designed according to the characteristics of each part. The overview of our unsupervised learning pipeline is shown in Fig. 1. For simplicity, the figure shows only the loss calculation in one direction ($t$ to $t-1$), and the loss calculation in the other direction ($t$ to $t+1$) is also used and is similar. Based on the image reconstruction with depth-pose or optical flow in Sec. III-A, the basic image reconstruction loss for depth-pose networks can be calculated from the input images $X_{t-1}$, $X_t$, the estimated depth map $D_t$ by the depth network, and the estimated pose transformation $T_{t-1}$ by the pose network; the basic image reconstruction loss for the optical flow network can be calculated from the input images $X_{t-1}$, $X_t$, and the estimated optical flow $f_{F(t-1)}$ by the optical flow network. Our proposed occlusion mask $M_{occ(t-1)}$ consisting of the edge mask $M_{edg(t-1)}$, the blank mask $M_{bla(t-1)}$ and the overlap mask $M_{ove(t-1)}$ is used to exclude the occlusion area in the image reconstruction losses. And the occlusion mask is computed without parameters in the process of bidirectional projection as in Sec. III-B.1. Our proposed less-than-mean masks $M_{ltm(t-1)}^{D-P}$ for depth and $M_{ltm(t-1)}^{F}$ for optical flow are used after the occlusion mask to exclude the interference of motion or illumination change in the image reconstruction losses as in Sec. III-B.2 and Sec. III-C.1. Our proposed maximum normalization method for the depth smoothness loss is used in the training of the depth-pose networks as in Sec. III-B.3. Our proposed consistency loss where the depth-pose flow $f_{D-P(t-1)}$ supervises the optical flow $f_{F(t-1)}$ in the occlusion area, is as in Sec. III-C.2. From the Fig. 1, it can be seen that we explicitly consider the occlusion problem to exclude the occlusion interference in the unsupervised training of depth-pose and optical flow, and design a new consistency loss for the training of optical flow in the occlusion area instead of simply segmenting the static and dynamic areas of an image as in CC [23]. What's more, the less-than-mean mask for the interference of motion and illumination change and the inverse depth maximum normalization for longer training also contribute to our pipeline.

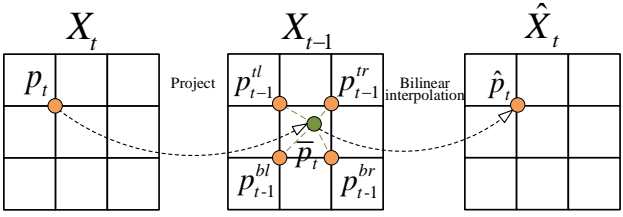

Fig. 2. Schematic diagram of the bilinear interpolation process. A pixel $p_t$ in the original image $X_t$ is projected to the coordinate position $\bar{p}_t$ on the adjacent frame $X_{t-1}$. The pixel value of the reconstructed $\hat{p}_t$ can be obtained by the pixel values of the nearest four points of $\bar{p}_t$.

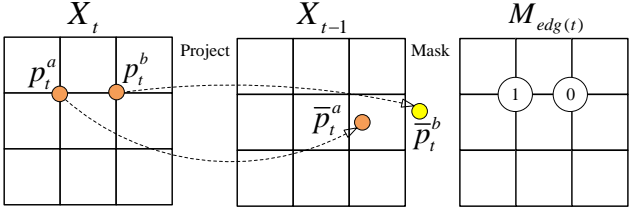

Fig. 3. Schematic diagram of the edge mask. The two pixels $p_t^a$ and $p_t^b$ in the original image $X_t$ are projected to $\bar{p}_t^a$ and $\bar{p}_t^b$ on the adjacent frame $X_{t-1}$. As $p_t^b$ is projected outside the image, the position at the point $p_t^b$ in the $M_{edg(t)}$ should be 0, and the position at $p_t^a$ is recorded as 1. The edge mask $M_{edg(t)}$ is obtained.

In this section, we first discuss the basic idea of unsupervised training using image reconstruction by depth-pose and by optical flow. Then we analyze the mismatch problem leading to regional distinctions of image. And it is separately discussed how to solve this problem by designing masks and loss functions in the training of depth-pose networks and the training of optical flow network.

*A. Image Reconstruction of Unsupervised Training of Depth-Pose Networks and Optical Flow Network*

In this paper, the key to unsupervised training of depth-pose networks and optical flow network is image reconstruction. For three consecutive frames $(X_{t-1}, X_t, X_{t+1})$, the depth $D_t$ of the intermediate frame $X_t$ can be obtained by depth net $N_{depth}$. The pose transformation $T_s$ between the intermediate frame and the adjacent frames can be obtained by pose net $N_{pose}$, where $s \in \{t-1, t+1\}$ indicates $t-1$ frame and $t+1$ frame. $T_s$ indicates the pose transformation of the camera from $t$ frame to $s$ frame.

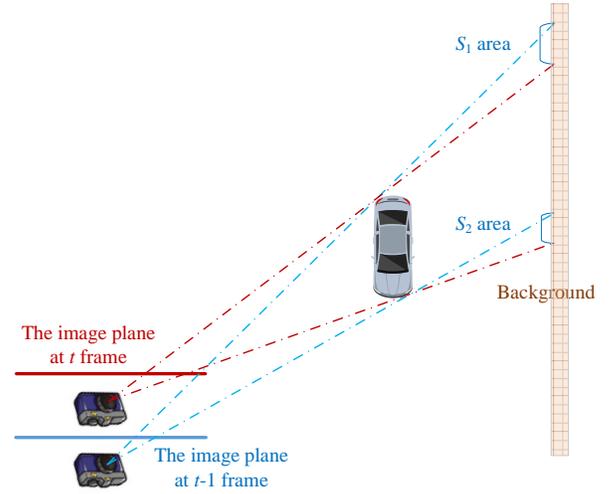

Fig. 4. The cause of the overlap mask. $S_1$ area can be seen at the $t$ frame but cannot be seen at the $t-1$ frame, which is because the $S_1$ area is blocked by the tail part of the car in the camera coordinate system of $t-1$ frame. Therefore, the pixels corresponding to the $S_1$ area in image $X_t$ will be mismatched with the tail portion of the car in $X_{t-1}$ when projected onto $X_{t-1}$, resulting in erroneous image reconstruction.

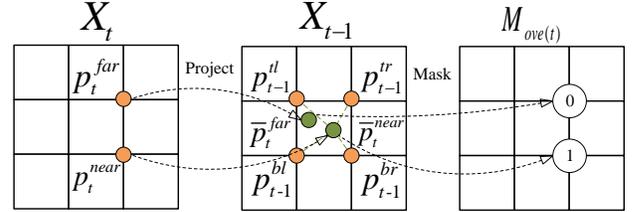

Fig. 5. Schematic diagram of the overlap mask. The two pixel $p_t^{far}$ and $p_t^{near}$ in the original image $X_t$ are projected to $\bar{p}_t^{far}$ and $\bar{p}_t^{near}$ on the adjacent frame $X_{t-1}$. $\bar{p}_t^{far}$ and $\bar{p}_t^{near}$ need to use the same values of the nearest four pixels for interpolation. At this time, the position at the farther point $p_t^{far}$ in the $M_{ove(t)}$ is recorded as 0, and the position at $p_t^{near}$ is recorded as 1. The overlap mask $M_{ove(t)}$ is obtained.

The point cloud $Q_t$ can be obtained for each pixel $(i, j)$ with estimated depth $D_t$ by $Q_t = D_t \cdot K^{-1}[i, j, 1]^T$, where $K$ is the camera intrinsic matrix.

Then, point cloud $\hat{Q}_s$ in the adjacent frame can be obtained by $\hat{Q}_s = T_s Q_t$. Afterward, the camera model can be used to project $\hat{Q}_s$ onto the image plane of the $s$ frame, denoted as $(\hat{i}, \hat{j})$, that is $[\hat{i}, \hat{j}, 1]^T = K\hat{Q}_s$. The formula for all relationship

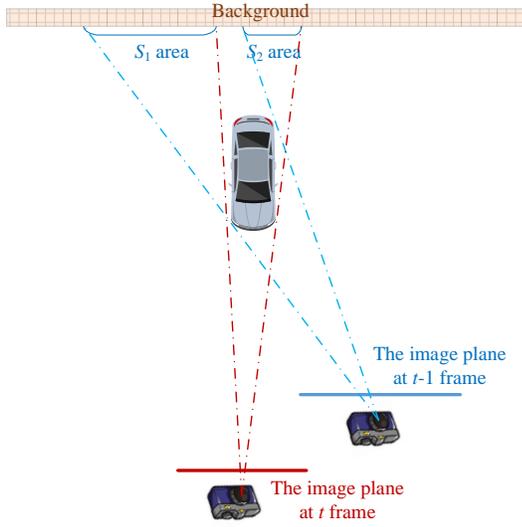

Fig. 6. The necessity of the blank mask. For $S_1$ area, there are similar problems as shown in Fig. 4. However, since the left side of the car cannot be seen in $X_t$, it is impossible to form near pixels as in Fig. 4 to block distant pixels to filter out pixels in the $S_1$ area. The back-projection is considered. When $X_{t-1}$ is projected onto $X_t$, since the $S_1$ region is not visible in the $t-1$ frame, a projection blank is generated in the region corresponding to $S_1$ on the $X_t$.

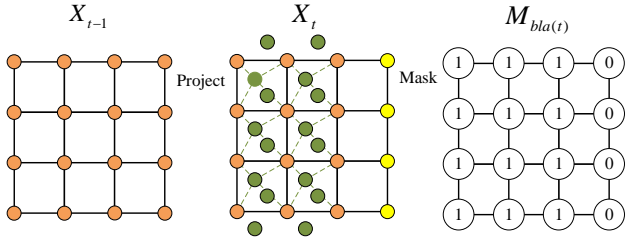

Fig. 7. Mechanism of the blank mask. All pixels in the original image $X_{t-1}$ are projected to the corresponding green dot positions on adjacent frames $X_t$. Now yellow dots in $X_t$ do not participate in interpolation, and the blank mask $M_{bla(t)}$ can be obtained.

above is written as:

$$[\hat{i}, \hat{j}, 1]^T = KT_s(D_s \cdot K^{-1}[i, j, 1]^T). \quad (1)$$

Then the reconstructed image of the intermediate frame can be obtained by differentiable bilinear interpolation [33], and the formula is $\hat{X}_s(p_t) = \sum_{i \in \{t, b\}, j \in \{l, r\}} w^{ij} X_s(p_s^{ij})$, $\sum_{i,j} w^{ij} = 1$, shown in Fig. 2, where $w^{ij}$ is the weighting factor of each pixel. The consistency between the reconstructed image and the original image is used to perform unsupervised training of depth-pose networks.

For optical flow network, $[\hat{i}, \hat{j}]^T = [i, j]^T + [u_s, v_s]^T$ can be directly obtained through the output $[u_s, v_s]^T$ of the optical flow network $N_{flow}$. Then, the bilinear interpolation method can also be used to reconstruct the image for unsupervised training.

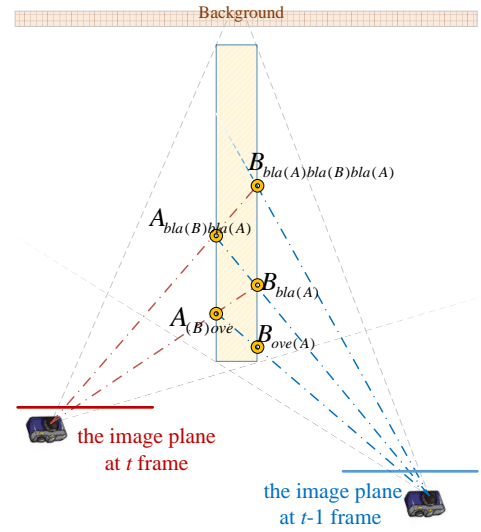

Fig. 8. Blank set corresponding to blank mask in the process of multiple mutual projection between $t$ frame and $t-1$ frame.

This paper refers to the basic image reconstruction loss of the reconstructed intermediate image and the original intermediate image as $L_R$. The formula is:

$$L_R = \sum_{s \in \{t-1, t+1\}} \sum_\Omega \left\| \sigma(X_t, \hat{X}_s) \right\|. \quad (2)$$

The function $\sigma(X_t, \hat{X}_s)$ is concerning the robust loss in [23], which includes the structural similarity (SSIM) in [34]. The formulas are:

$$\sigma(x, y) = \lambda_\rho \sqrt{(x-y)^2 + \varepsilon^2} + (1-\lambda_\rho) SSIM(x, y), \quad (3)$$

$$SSIM(x, y) = \left[ 1 - \frac{(2\mu_x\mu_y + c_1)(2\mu_{xy} + c_2)}{(\mu_x^2 + \mu_y^2 + c_1)(\sigma_x + \sigma_y + c_2)} \right], \quad (4)$$

Where $\lambda_\rho = 0.15$, $\varepsilon = 0.01$, $c_1 = 0.01^2$, $c_2 = 0.03^2$. But this loss function requires a correct pixel match between the reconstructed image and the original image, which cannot be achieved sometimes in the unsupervised training of depth-pose networks as well as the optical flow network. We first solve the problem of the depth-pose networks in Sec. III-B. Then in Sec. III-C we introduce the idea into the training of optical flow that depth and pose are used to calculate the occlusion mask, and introduce a new unsupervised loss in optical flow training.

### B. Unsupervised Training of Depth-Pose Networks
#### 1) Edge Mask, Overlap Mask and Blank Mask for Occlusion

The accuracy of image reconstruction directly affects the direction of unsupervised training, and occlusion is a factor that significantly affects image reconstruction. The occlusion that some pixels in the first frame are projected outside the image of the adjacent frame has been discussed in [20] and [21], defined as edge mask and symbolized as $M_{edg}$ in this paper. Schematic diagram of the edge mask is as shown in Fig. 3.

However, this is only a part of the occlusion. The example analysis of one situation in which the edge mask cannot solve is shown in Fig. 4. Here we add an overlap mask to solve this problem. As shown in Fig. 5, when two pixels in $X_t$ are projected into the same pixel grid of $X_{t-1}$, The farther pixel $\bar{p}_t^{far}$ is judged as occlusion. The nearer pixel $\bar{p}_t^{near}$ is judged as non-occlusion. This creates an overlap mask, denoted as $M_{ove(t)}$.

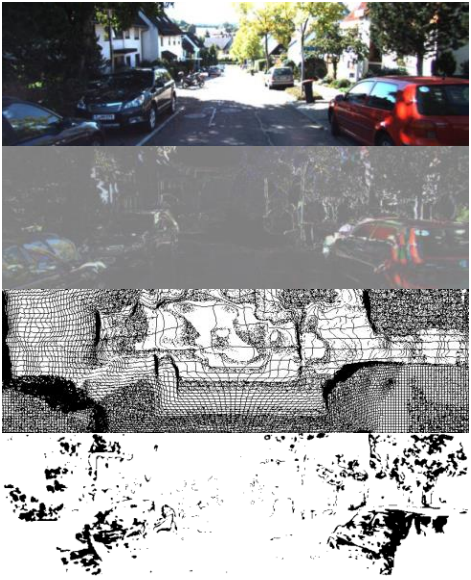

Fig. 9. Visualization comparison of occlusion mask and less-than-mean mask. Top to bottom: input image, difference image between raw image and reconstructed image, occlusion mask and less-than-mean mask.

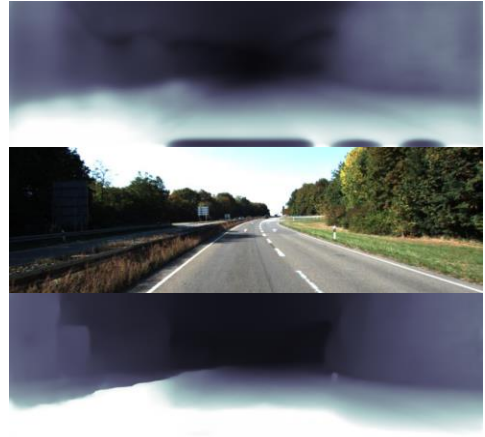

Fig. 10. Visualization comparison of two normalization effects. Top to bottom: depth estimation map trained with mean normalization to 14100th generation, input image and depth estimation map trained with maximum normalization to 14100th generation. The deeper the color is, the greater the depth value is. The models for both normalizations are initialized and trained with the same weights and the same number of generations.

However, this does not solve all the problems. Fig. 6 shows an example that the overlap mask cannot handle. But this problem can be solved by back projection. As shown in Fig. 7, when the pixels in the $X_{t-1}$ frame are projected onto the $X_t$, a projection blank region is generated and some pixels in $X_t$ do not participate in the interpolation calculation. From these pixels that do not participate in the interpolation, we can obtain the blank mask $M_{bla(t)}$.

Can the edge mask, overlap mask, and blank mask solve the mismatched pixels caused by camera ego-motion during image reconstruction? We analyze the question from Set Theory.

**Definition 1.** Define the pixels in $X_t$ as set $A$, pixels in $X_{t-1}$ as set $B$. Pixels in set $A$ (or $B$) are projected into 3D space. If the $t-1$ frame (or $t$ frame) camera cannot catch a 3D space point corresponding to a certain pixel in $A$ (or $B$), the pixel is occluded and it should be masked. Otherwise, the pixel also belongs to $B$ (or $A$).

**Proposition 1.** There are two reasons why the pixels in the set $A$ are not visible on the image plane of the $t-1$ frame. On the one side, they are projected outside the pixel plane of the $t-1$ frame. On the other side, the pixel is blocked by other pixels at the $t-1$ frame. Pixels projected outside the $X_{t-1}$ image are recorded as the set $A_{edg}$ ($A_{edg}$ belongs to $A$ but does not belong to $B$). At the camera viewpoint of $t-1$ frame, part of the pixels in $A$ which belongs to set $A_{(B)}$ (Element belongs to $A$ and belongs to $B$ at the same time) occludes another set of pixels, recorded as the set $A_{(A)ove}$, leading $A_{(A)ove}$ to be invisible under $t-1$ frames and should be masked. However, some pixels in $A$ denoted as set $A_{(B)ove}$ are also occluded by other pixels and the pixels that cause occlusion are not in the set $A$ (That is, not belonging to the set $A$, because these pixels are occluded under $t-1$ frames). So these pixels must belong to set $B$, that is, $B_{ove(A)}$ occludes $A_{(B)ove}$ in the space. Then, when all the pixels in the set $B$ are projected onto $X_{t-1}$, since no pixels correspond to $A_{(B)ove}$ (Because $B_{ove(A)}$ in set $B$ is in front of $A_{(B)ove}$, the camera does not catch $A_{(B)ove}$ in the $t-1$ frame), the region corresponding to $A_{(B)ove}$ will be blank, which means $A_{(B)ove}$ should be masked. This is the blank mask discussed before.

But there is a rarer situation as shown in Fig. 8. Although the set $B_{ove(A)}$ occludes the set $A_{(B)ove}$ from the perspective of $t-1$ frame, there are still some pixels $B_{bla(A)}$ projected in the $A_{(B)ove}$ so that the region corresponding to $A_{(B)ove}$ cannot completely be blank in the $t$ frame. However, when the $A$ set is projected to the $t-1$ frame, the corresponding region of $B_{bla(A)}$ should be blank. This is because $B_{bla(A)}$ is projected to the $A_{(B)ove}$ region in the $t$ frame, and $A_{(B)ove}$ is already in set $A$, so $B_{bla(A)}$ must be occluded by $A_{(B)ove}$. If the $B_{bla(A)}$ region does not produce a blank when the pixels in the $t$ frame are projected to the $t-1$ frame, it means that $A_{bla(B)bla(A)}$ in the set $A$ can also be projected to the $B_{bla(A)}$. But in the $t-1$ frame, $A_{bla(B)bla(A)}$ is not visible, indicating that there will be blank in the $A_{bla(B)bla(A)}$ region when the pixels in $t-1$ frame are projected to the $t$ frame. If the position $A_{bla(B)bla(A)}$ does not produce a blank because of $B_{bla(A)bla(B)bla(A)} \cdots$. From this continuous reasoning, it can be indicated that there might always be new occlusion pixels that should be blank. However, since the number of pixels in the two sets is limited, finally there is no pixel to break the blanking assumption at the next step. Therefore, by continuously mutual projecting, all the mismatched pixels because of occlusion in the two sets can be removed finally.

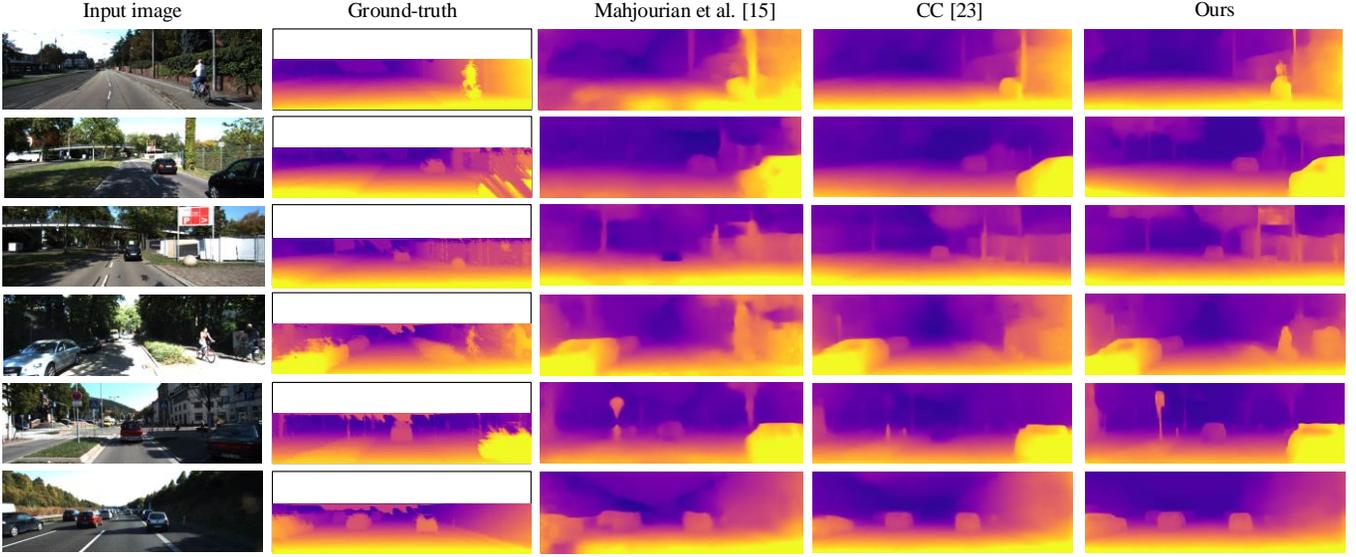

Fig. 11: The visual comparison of depth estimation among Mahjourian et al. [15], CC [23] and ours on Eigen test set [7]. The depth visualization of ground truth are obtained by interpolation.

**Remark 1.** A single forward-backward projection can remove most of the occluded pixels and multiple projections will increase the amount of calculation, but the actual effect is low, so that only one mutual projection is executed for the experiments in this paper to get the edge, overlap and blank masks. Then these three masks are multiplied up to get the occlusion mask $M_{occ}$. The formula is:

$$M_{occ(s)} = M_{edg(s)} M_{ove(s)} M_{bla(s)}. \quad (5)$$

*2) Less-than-mean Mask and Image Reconstruction Loss for the Training of Depth-Pose Networks*

A few occluded, motion and illumination change regions remain in the process of image reconstruction using depth-pose networks. When calculating the image reconstruction loss, these outliers do not satisfy the image matching relationship, which causes a disturbance to the reconstruction loss calculation. In order to eliminate pixels that have too large reconstruction error values, all outliers larger than the average image reconstruction loss are masked by a less-than-mean mask $M_{ltm(s)}^{D-P}$. The formula is:

$$M_{ltm(s)}^{D-P} = \mathbb{I}\left(\sigma(X_t, \hat{X}_s) < \frac{1}{\Omega}\sum_\Omega \left\|\sigma(X_t, \hat{X}_s) \cdot M_{occ(s)}\right\|\right), \quad (6)$$

where $s \in \{t-1, t+1\}$. $\mathbb{I}(\cdot)$ is an indicator function. Once the equation in the parenthesis is satisfied, the corresponding value equals 1. As shown in Fig. 9, this can further remove the disturbance brought by the outliers and makes the training data more effective. Therefore, final image reconstruction loss of depth-pose unsupervised training is:

$$L_{R(D-P)} = \sum_{s\in\{t-1,\,t+1\}} \sum_\Omega \left\|\sigma(X_t, \hat{X}_s) M_{occ(s)} M_{ltm(s)}^{D-P}\right\|. \quad (7)$$

It can be seen clearly from the difference image in Fig. 9 that because of motion, illumination change, etc., the reconstruction loss of the car body is larger. Less-than-mean mask calculates the average reconstruction loss of the whole image after masking occlusion area and excludes illumination change area whose loss is larger than average, which shows that less-than-mean mask further eliminates outliers for training.

*3) Smoothness Loss for Depth-Pose Training*

Smoothness loss can make the estimated depth map $(D)$ smoother and reduces the disturbance from outliers. In the initial study of monocular depth and pose estimation, the output inverse depth, disparity $(D^{-1})$, of the depth network was directly used to calculate the smoothness loss, which makes the inverse depth smoother. The basic formula is:

$$L_{S(D-P)} = \sum_\Omega \left\|\nabla(D^{-1}) \cdot e^{-\nabla X}\right\|^2. \quad (8)$$

But from the formula, we can find that the smaller the value of $D^{-1}$, the smaller the gradient $\nabla(D^{-1})$, which causes that at the end of the training, all the values in the inverse depth map are decreasing and approaching 0, making the training invalid. Wang et al. [16] proposed a normalization method that the inverse depth map is divided by $(D^{-1})_{mean}$, the mean of all inverse depth values. And when calculating smoothness loss, the input is $X = \dfrac{D^{-1}}{(D^{-1})_{mean}}$. The principle is reducing the sensibility of smoothness loss for absolute scale, so as to avoid the problem of the inverse depth scale reduction in the later

TABLE I
COMPARISON FOR THE DEPTH ESTIMATION AMONG OTHER MONOCULAR DEPTH ESTIMATION METHODS AND OURS TESTED ON EIGEN ET AL. [7] TEST SPLIT

| Method | Supervised | Dataset | Error metric (lower are better) | | | | Accuracy metric (higher are better) | | |
|---|---|---|---|---|---|---|---|---|---|
| | | | Abs Rel | Sq Rel | RMSE | RMSE log | $\delta < 1.25$ | $\delta < 1.25^2$ | $\delta < 1.25^3$ |
| Eigen et al. [7] Coarse | Depth | K | 0.214 | 1.605 | 6.563 | 0.292 | 0.673 | 0.884 | 0.957 |
| Eigen et al. [7] Fine | Depth | K | 0.203 | 1.548 | 6.307 | 0.282 | 0.702 | 0.890 | 0.958 |
| Liu et al. [8] | Depth | K | 0.202 | 1.614 | 6.523 | 0.275 | 0.678 | 0.895 | 0.965 |
| Godard et al. [11] | Stereo | K | 0.148 | 1.344 | 5.927 | 0.247 | 0.803 | 0.922 | 0.964 |
| Zhan et al. [12] | Stereo | K | 0.144 | 1.391 | 5.869 | 0.241 | 0.803 | 0.928 | 0.969 |
| Zhou et al. [14] | No | K | 0.208 | 1.768 | 6.856 | 0.283 | 0.678 | 0.885 | 0.957 |
| LEGO [20] | No | K | 0.162 | 1.352 | 6.276 | 0.252 | - | - | - |
| Mahjourian et al. [15] | No | K | 0.163 | 1.240 | 6.220 | 0.250 | 0.762 | 0.916 | 0.968 |
| GeoNet VGG [21] | No | K | 0.164 | 1.303 | 6.090 | 0.247 | 0.765 | 0.919 | 0.968 |
| GeoNet ResNet [21] | No | K | 0.155 | 1.296 | 5.857 | 0.233 | 0.793 | 0.931 | 0.973 |
| Shen et al. [18] | No | K | 0.156 | 1.309 | 5.73 | 0.236 | 0.797 | 0.929 | 0.969 |
| Wang et al. [19] | No | K | 0.154 | 1.163 | 5.700 | 0.229 | 0.792 | 0.932 | 0.974 |
| DF-Net [22] | No | K | 0.150 | 1.124 | 5.507 | 0.223 | 0.806 | 0.933 | 0.973 |
| GANVO [17] | No | K | 0.150 | 1.141 | 5.448 | **0.216** | 0.808 | 0.939 | 0.975 |
| CC [23] | No | K | **0.140** | 1.070 | 5.326 | 0.217 | 0.826 | 0.941 | 0.975 |
| Ours | No | K | **0.140** | **1.068** | **5.255** | 0.217 | **0.827** | **0.943** | **0.977** |
| Godard et al. [11] | Stereo | CS + K | **0.124** | 1.076 | 5.311 | 0.219 | **0.847** | 0.942 | 0.973 |
| Zhou et al. [14] | No | CS + K | 0.198 | 1.836 | 6.565 | 0.275 | 0.718 | 0.901 | 0.960 |
| LEGO [20] | No | CS + K | 0.159 | 1.345 | 6.254 | 0.247 | - | - | - |
| Mahjourian et al. [15] | No | CS + K | 0.159 | 1.231 | 5.912 | 0.243 | 0.784 | 0.923 | 0.970 |
| GeoNet ResNet [21] | No | CS + K | 0.153 | 1.328 | 5.737 | 0.232 | 0.802 | 0.934 | 0.972 |
| Shen et al. [18] | No | CS + K | 0.152 | 1.205 | 5.564 | 0.227 | 0.8 | 0.935 | 0.973 |
| Wang et al. [19] | No | CS + K | 0.155 | 1.184 | 5.765 | 0.229 | 0.790 | 0.933 | 0.975 |
| DF-Net [22] | No | CS + K | 0.146 | 1.182 | 5.215 | 0.213 | 0.818 | 0.943 | **0.978** |
| GANVO [17] | No | CS + K | 0.138 | 1.155 | 4.412 | 0.232 | 0.820 | 0.939 | 0.976 |
| CC [23] | No | CS + K | 0.139 | 1.032 | 5.199 | 0.213 | 0.827 | 0.943 | 0.977 |
| Ours | No | CS + K | 0.132 | **0.986** | **5.173** | **0.212** | 0.835 | **0.945** | 0.977 |

Note: The maximum depth value for the evaluation is 80m. 'CS+K' means firstly pre-training on Cityscapes dataset [31], then training on the KITTI dataset [9]. 'K' means training on the KITTI dataset [9] only. The best results are bold in each block.

TABLE II
ABLATION STUDY RESULTS OF DEPTH ESTIMATION TESTED ON EIGEN ET AL. [7] TEST SPLIT

| Method | Supervised | Dataset | Error metric (lower are better) | | | | Accuracy metric (higher are better) | | |
|---|---|---|---|---|---|---|---|---|---|
| | | | Abs Rel | Sq Rel | RMSE | RMSE log | $\delta < 1.25$ | $\delta < 1.25^2$ | $\delta < 1.25^3$ |
| Basic | No | K | 0.145 | 1.098 | 5.527 | 0.226 | 0.817 | 0.936 | 0.973 |
| Basic + Occ | No | K | 0.144 | 1.121 | 5.510 | 0.223 | 0.818 | 0.939 | 0.974 |
| Basic + Occ + LTM | No | K | 0.142 | 1.090 | 5.423 | 0.218 | 0.822 | 0.940 | 0.976 |
| Basic + Occ + LTM+ MN | No | K | **0.140** | **1.068** | **5.255** | **0.217** | **0.827** | **0.943** | **0.977** |

Note: 'Basic' means baseline setting. 'Occ' means occlusion mask. 'LTM' means less-than-mean mask. 'MN' means maximum normalization.

stage of training.

But our experiments demonstrated that in the later stage of training with this normalization, the inverse depth is still approaching 0 for textureless areas and illumination change areas. After the analysis, we demonstrated that since the reconstructed pixel cannot match the target frame, image reconstruction loss in these pixels will not decrease with training, and smoothness loss will become dominant in these areas. For local areas where smoothness loss is dominant, the change of $(D^{-1})_{mean}$ for the full image is relatively small, which is considered to be unchanged. Thus, the smaller the inverse depth value of these local regions, the smaller the smoothness loss. Therefore, in the later stage of training, the inverse depth of the regions where the reconstruction error is relatively large will approach 0, that is the depth tends to infinity.

In order to solve this problem, we propose to divide the inverse depth $D^{-1}$ of each pixel by the maximum $(D^{-1})_{max}$ of all the inverse depth values before normalization. Then smoothness loss is calculated by using the reciprocal of the normalized inverse depth $X = \frac{(D^{-1})_{max}}{D^{-1}}$. This can directly solve the above problems. The analysis is as follows.

For the input of the smoothness loss function, make the following derivation:

$$X = \frac{(D^{-1})_{max}}{D^{-1}} = \frac{\frac{1}{D_{min}}}{\frac{1}{D}} = \frac{D}{D_{min}}. \quad (9)$$

The domain of $\frac{D}{D_{min}}$ is always $[1, +\infty)$. Pixels with large reconstruction errors are constrained by $D_{min}$ when calculating

TABLE III
ABSOLUTE TRAJECTORY ERROR (ATE) COMPARED WITH THE RECENT WORKS ON POSE ESTIMATION

| Method | Seq.09 | Seq.10 |
|---|---|---|
| ORB-SLAM (full) | 0.014 ± 0.008 | 0.012 ± 0.011 |
| ORB-SLAM (short) | 0.064 ± 0.141 | 0.064 ± 0.130 |
| Zhou et al. [14] | 0.021 ± 0.017 | 0.020 ± 0.015 |
| Mahjourian et al. [15] | 0.013 ± 0.010 | 0.012 ± 0.011 |
| GeoNet [21] | 0.012 ± 0.007 | 0.012 ± 0.009 |
| CC [23] | 0.012 ± 0.007 | 0.012 ± 0.008 |
| GANVO [17] | 0.009 ± 0.005 | 0.010 ± 0.013 |
| Wang et al. [19] | 0.009 ± 0.005 | 0.008 ± 0.007 |
| Shen et al. [18] | 0.0089 ± 0.0054 | 0.0084 ± 0.0071 |
| Ours | **0.0081 ± 0.0044** | **0.0082 ± 0.0064** |

Note: The lower the value is, the better the result is.

TABLE IV
COMPARISON WITH THE RECENT WORKS ON OPTICAL FLOW ESTIMATION

| Method | Train | | Test |
|---|---|---|---|
| | EPE | F1 | F1 |
| FlowNet2 [25] | 10.06 | 30.37% | - |
| SPyNet [26] | 20.56 | 44.78% | - |
| UnFlow-C [28] | 8.80 | 28.94% | 29.46% |
| UnFlow-CSS [28] | 8.10 | **23.27%** | - |
| Back2Future [30] | **6.59** | - | 22.94%* |
| Geonet [21] | 10.81 | - | - |
| DF-Net [22] | 8.98 | 26.01% | 25.70% |
| CC [23] | **6.21** | 26.41% | - |
| Ours | 6.66 | **23.04%** | **22.36%*** |

Note: 'EPE' means average endpoint error, 'F1' means the percentage of pixels whose EPE is greater than 3 pixels and greater than 5% of the ground truth. The lower 'EPE' and 'F1' values are better. '*' means that the results are obtained after supervised fine-tune and tested on the Optical Flow 2015 test set [9].

TABLE V
ABLATION STUDY RESULTS OF OPTICAL FLOW ESTIMATION

| Method | Train | | Test |
|---|---|---|---|
| | EPE | F1 | F1 |
| Basic | 12.61 | 28.06% | - |
| Basic + Occ | 11.51 | 27.56% | - |
| Basic + Occ + LTM | 11.35 | 27.10% | - |
| Basic + Occ + LTM + Cons | 6.66 | 23.04% | - |

Note: Ablation studies for optical flow estimation. 'Basic' means baseline setting. 'Occ' means occlusion mask. 'LTM' means less-than-mean mask. 'Cons' means consistency loss.

the smoothness loss. Simply reducing these depth values to zero can't reduce the smoothness loss, thus avoiding the depth of this partial region approaching zero in training.

As shown in Fig. 10, when using the mean normalization to deal with the inverse depth in the smoothness loss, the reconstruction loss of the near road area cannot be reduced, and the smoothness loss tends to make the depth infinity in 14100th generation. As the depth map at the top shows, the area turns black. And our maximum normalization smoothness loss can directly solve such problems, as shown in the depth map in the bottom, the image is still normal when training to 14100th generations.

### C. Unsupervised Training of Optical Flow Assisted by Depth and Pose

#### 1) Image Reconstruction Loss and Smoothness Loss for Unsupervised Training of Optical Flow

The image reconstruction loss of optical flow is also invalid in the occluded region, and mismatched pixels will mislead the training. Since the occluded regions based on the depth-pose networks have already been calculated explicitly, the occlusion mask calculated by formula (5) is directly adopted to remove mismatched pixels in optical flow image reconstruction. Then like in Sec. III-B.2, the less-than-mean mask is used to further remove the outliers. Specifically, after masking the occluded regions, the pixels whose reconstruction errors are higher than the average error of the whole image are masked. The mask is denoted as $M^F_{ltm(s)}$. Combined with the basic formula (2), the image reconstruction loss for optical flow is:

$$L_{R(F)} = \sum_{s \in \{t-1, t+1\}} \sum_{\Omega} \left\| \sigma(X_t, \hat{X}_s) M_{occ(s)} M^F_{ltm(s)} \right\|. \quad (10)$$

The smoothing loss of optical flow is the same as that of the previous unsupervised work of optical flow [21-23]. The gradient of the image is used to add a prior constraint to the gradient of the optical flow. The optical flow output of the optical flow network is denoted as $f_{F(s)}$. The formula for the smoothing loss of optical flow is as follows:

$$L_{S(F)} = \sum_{s \in \{t-1, t+1\}} \sum_{\Omega} \left\| \nabla(f_{F(s)}) \cdot e^{-\nabla X} \right\|^2. \quad (11)$$

#### 2) Consistency Loss

Occluded region cannot be trained in image reconstruction loss of optical flow, while depth and pose can obtain depth-pose flow in occluded regions. According to Wang et al. [35], since in most cases occluded regions are rigid regions, rigid flow obtained by depth and pose could be used to supervise optical flow in occluded regions. As shown in Fig. 9, the full occlusion mask includes additional stripes which are generated from an overlap mask. The pixels masked by these stripes are not involved in the image reconstruction loss, which will not mislead the training of optical flow networks. However, in the region of these pixels, the training of optical flow will be misled if depth-pose flow is used to supervise the optical flow, as many of these pixels do not belong to the occluded region and even belong to a dynamic region where consistency between optical flow and depth-pose flow is not satisfied. Therefore, we use the edge and blank masks here. The formula for the consistency loss of optical flow is as follows:

$$L_C = \sum_{s \in \{t-1, t+1\}} \sum_{\Omega} \left\| \sigma(f_{D-P(s)}, f_{F(s)}) \cdot (1 - M_{edg(s)} M_{bla(s)}) \cdot M^F_{ltm(s)} \right\|, \quad (12)$$

where rigid flow $f_{D-P(s)} = [\hat{i}, \hat{j}]^T - [i, j]^T$ can be obtained directly from the projection formula (1).

## IV. EXPERIMENT AND EVALUATION

The experiments are mainly carried out on the KITTI dataset [9]. As many previous works report the depth estimation results obtained by training on KITTI [9] after pre-training on

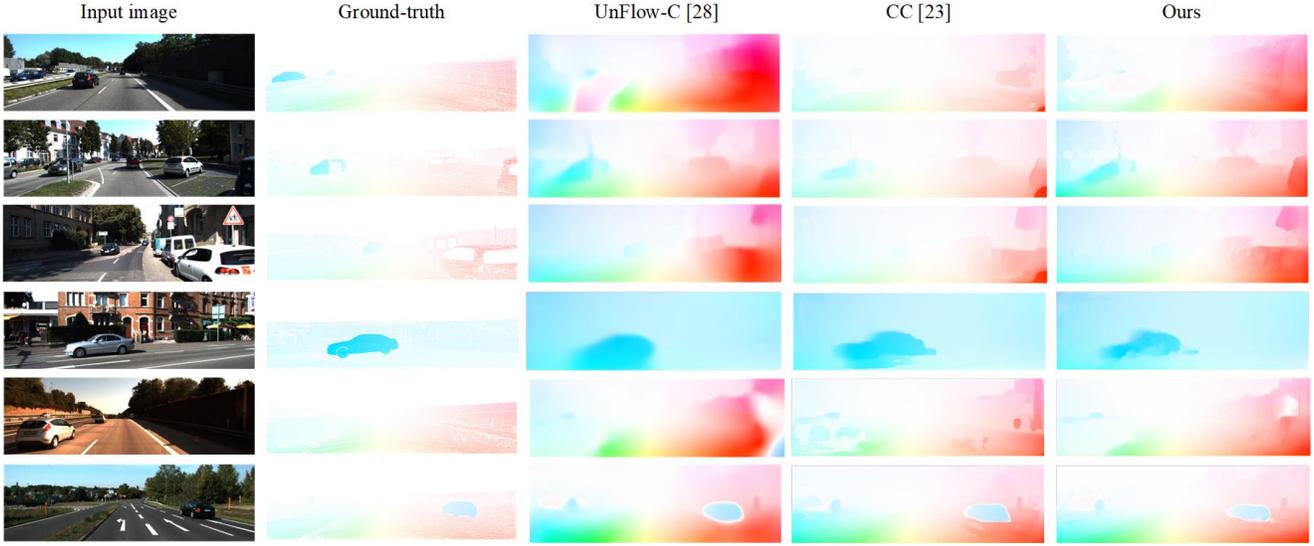

Fig. 12: Comparison of flow estimations on KITTI 2015 flow datasets [9] among UnFlow-C [28], CC [23] and ours.

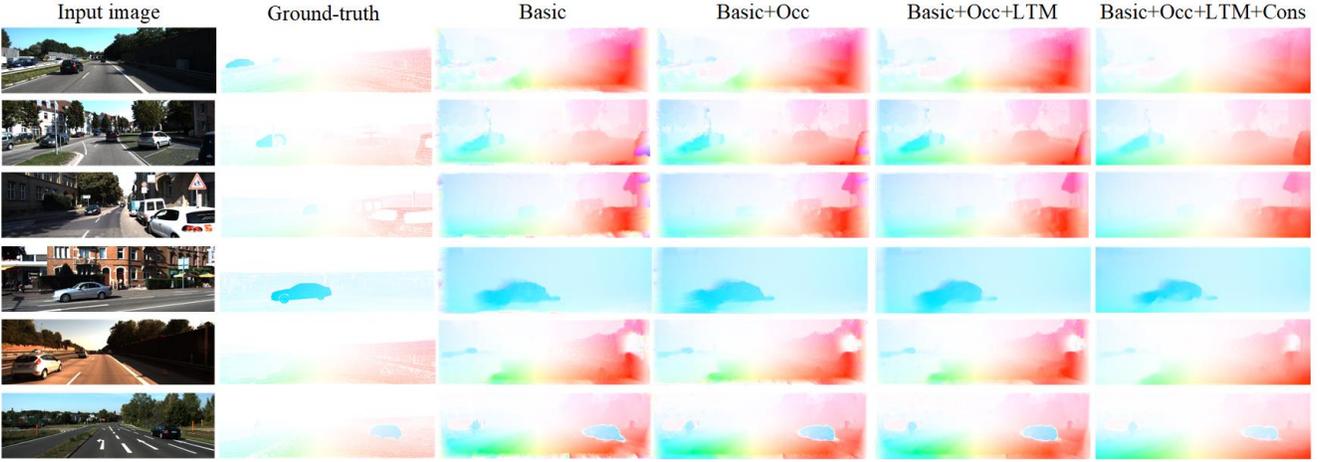

Fig. 13: Comparison of flow estimations on KITTI 2015 flow datasets [9] among our ablation studies. It is shown that the results of flow estimation are improved by our treatment for occlusion, outliers and our consistency loss.

Cityscape dataset [31], we also add this part.

*A. Training Settings*

Only one-scale images are used in the unsupervised training of depth and pose, which is different from the training of optical flow. In the unsupervised training process of the optical flow, six scales $l$ of images are used. The total loss function for the unsupervised training of depth and pose is given by:

$$L_{D-P} = w_{R(D-P)}L_{R(D-P)} + w_{S(D-P)}L_{S(D-P)}, \quad (13)$$

where $\{w_{R(D-P)}, w_{S(D-P)}\}$ are weights of the individual loss functions and $w_{R(D-P)} = 1.0$, $w_{S(D-P)} = 0.2$.

The total loss function for the unsupervised training of optical flow model is given by:

$$L_{Flow} = \sum_{l}(w_{R(F)}L_{R(F)}^{l} + w_C L_C^l + w_{S(F)}L_{S(F)}^l), \quad (14)$$

where $\{w_{R(F)}, w_C, w_{S(F)}\}$ are weights of the individual loss functions and $w_{R(F)} = 1.0$, $w_C = 0.1$, $w_{S(F)} = 0.005$.

The network architecture in CC [23] is adopted by ours, and its optical flow network refers to PWC-Net [36] while the depth network refers to ResNet [37].

The KITTI dataset [9] and the Cityscape dataset [31] are used for training. The KITTI dataset is a commonly used computer vision algorithm evaluation dataset for autonomous driving scenarios which contains real image data collected from urban, rural, and highway scenes. The Cityscapes dataset is a new large-scale dataset that contains a set of different stereo video sequences recorded in street scenes in 50 different cities. 33657 and 11055 samples from KITTI are used as training and validation sets. 62049 samples from Cityscape are used as the pre-training set. Each sample contains three consecutive frames. All images are scaled to 256 × 832. Random scaling, cropping, and horizontal flips are adopted for data augmentation.

The experiments are performed on a 2080Ti GPU. The training process continues until the validation error is saturated.

For better results, we first train the depth and pose network, using our maximum normalization, the calculated occlusion mask and less-than-mean mask to get the trained depth and pose models. First, the models are pre-trained on the Cityscape dataset, which spent 20 hours for 40000 steps. Then, the models get trained on KITTI and the training process lasted about 70 hours for 149850 steps. Next, we use the trained depth and pose models in the training process of optical flow, in which the occlusion region calculated by the depth and pose models is not involved in the training of optical flow. Besides, in the region masked by the edge and blank masks, rigid flow calculated by depth and pose is used to supervise the training of optical flow. The training process of the optical flow network lasted about 90 hours for 190K steps. We use Adam optimizer [38] with 4 as batch size [39] for the depth-pose phase and 12 as batch size for the optical flow phase, and the learning rate is always $10^{-4}$.

### B. Depth Evaluation Results

Our depth evaluation process and evaluation metrics are consistent with the previous work [21]-[23]. Our depth evaluation uses the test set of Eigen et al. [7] test split. 697 images are involved in the evaluation. The evaluation process and the evaluation metrics are the same for all depth evaluation in this paper including the ablation study. The comparison results of depth estimation between ours and others are listed in Table I. [11] and [12] used binocular video which means extra pose supervision because of the constant pose between binocular. Compared to [7] and [8], our method doesn't need depth supervision. Compared to [15], complex ICP back-propagation is not adopted in our method. Compared to [20], our method does not need the aid of an edge network. Compared to [23], our method does not need an extra motion segmentation network and does not need a complex training scheme based on iteration. As shown in the Table I, our method outperforms most unsupervised methods. And our methods can even obtain competitive results with binocular supervised method [11]. We believe that it is because of our elimination of the mismatched pixels in image reconstruction and our additional designs for robustness.

As shown in Table II, we first set the baseline, that is, the image reconstruction loss and the smoothing loss are involved for the entire image. Then the occlusion mask, the less-than-mean mask, and the maximum normalization are involved successively as improvements. The training conditions are the same as the training setting in Sec. IV-A except for the experimental variables described in Table II and the pretraining is not used here. Each configuration is trained from scratch. All evaluations are on the Eigen et al. [7] test split. The evaluation process and the evaluation metrics are the same as Table I. The results show that each of our designs ameliorates the results.

The "Basic" takes 118881 training steps and about 40 hours. The "Basic + Occ" takes 101898 training steps and about 64 hours. The "Basic + Occ + LTM" takes 103896 training steps and about 65 hours. The "Basic + Occ + LTM + MN" takes 186813 training steps and about 104 hours. It can be seen that the training steps of the first three are similar, but the fourth one obviously increases. This is due to our normalization method. The training can be continued without the phenomenon of depth value degradation, resulting in continued growth of accuracy.

The visual comparison of the depth estimation results is shown in Fig. 11. Our depth estimation performs better on dynamic objects such as moving persons and cars. The estimations of distant objects are also more accurate, like further cars and trees. Our method can provide more details as can be observed in the traffic sign shape. The advantages are obtained by using spatial 3D geometry to mask occlusion and using the less-than-mean mask to eliminate misleading in training which occurs because of the illumination change and the mismatch on dynamic objects. In addition, smoothness loss with maximum normalization is proposed so that the training can last longer while degradation will not occur in textureless regions such as roads.

### C. Pose Evaluation Results

According to the general method [14], the camera pose is evaluated on the official KITTI odometry dataset. ORB-SLAM [4] and other recent state-of-the-art unsupervised learning results are compared in Table III. ORB-SLAM (full) has closed-loop and relocation while ORB-SLAM (short) does not. Table III shows that in the unsupervised monocular methods, our method calculating over 3-frame snippets achieves state-of-the-art performance which exceeds ORB-SLAM (full). We believe that in the training process, the rich mask information can assist the pose network to learn how to ignore the occlusion area for pose estimation.

### D. Optical Evaluation Results

Our optical flow evaluation process and evaluation metrics are consistent with the previous work [21]-[23], [28]. As shown in Table IV, we quantitatively compare our method with the CC [23] and other methods for optical flow estimation in KITTI 2015 training set [9] like previous work [21]-[23], [28]. 200 samples in the KITTI 2015 training set [9] are used for the evaluation of the optical flow model. The evaluation process is the same for all optical flow evaluations in this paper including the ablation study. Note that 3 cascaded networks are adopted in UnFlow-CSS [28] to refine optical flow. The method of splitting dynamic and static is adopted in CC [23], so that in the evaluation stage, the optical flow is evaluated by combining the depth network, the pose network, the optical flow network and the mask network. Our method, which only needs a trained optical flow network, can perform close to [23] in terms of average end point and much better than [23] in terms of outlier error. Our method also outperforms [21], [22], which use a larger ResNet-50 network.

This process benefits from our accurately calculated occlusion mask. When training the optical flow model, the image reconstruction error is not considered in the occlusion area, but depth and pose are used to supervise the optical flow training in the occlusion area instead. This method allows the optical flow network to learn how to estimate more accurately in occluded area.

The qualitative results are shown in Fig. 12. Our method performs better in occluded regions, which benefits from our occlusion model based on 3D geometry. Compared to [23], our flow estimation is smoother, which benefits from the usage of only one optical flow network instead of combining depth-pose flow and optical flow from 4 networks.

The ablation experiments are shown in Table V. The training conditions are the same as the training setting in Sec. IV-A except for the experimental variables described in Table II and the best depth-pose model is used here. Each configuration for the optical flow phase is trained from scratch based on the trained depth-pose model. All evaluation is on KITTI 2015 training set [9]. The evaluation process and the evaluation metrics are the same as Table IV. Maximum number of training steps is adopted as the stopping criterion for the ablation study of optical flow, because we found that there is no obvious over-fitting even if the training of optical flow reaches 190K steps. However, it is already saturated. So, we make all training of optical flow phase last about 90 hours and about 190K steps.

As shown in Table V, experiments show that each of our designs makes the results better, whose qualitative results are shown in Fig. 13. It is shown that there are many wrong optical flow values estimated in the edges of the basic group, that is because many of the edges are occluded regions and the optical flow network will be misled by the wrong image reconstruction in the occluded regions. To reduce errors in the occluded regions, corresponding masks are introduced from our depth-pose models. However, optical flows tend to be small for the "Basic + Occ" and "Basic + Occ + LTM" groups as compensatory strategy for optical flow estimation in occluded regions is not learned during the training. With the consistency loss in the "Basic + Occ + LTM + MN" group, depth and pose provide the strategy to the optical flow estimation in occluded regions, which markedly improves the estimation results. It also benefits from our depth and pose estimation.

## V. CONCLUSION

The tasks of depth, pose and optical flow have close correlation. In this paper, we make use of the correlation of three tasks, so that the trained depth and pose model can assist the optical flow training, and the performance of optical flow is also improved. Our method does not require repeated iterative training [23] but can get better depth estimation, pose estimation and optical flow estimation results. Contrast experiments with others demonstrate the superiority of our approach.

The use of rigid flow computed from the depth and pose to supervise optical flow in the occluded area is an approximate method, because the occluded area may not meet the rigidity assumption. The occluded area might contain dynamic objects. Fortunately, this situation is relatively rare in the entire image. More importantly, the supervised estimation with small error is better than no supervision. Therefore, supervising the optical flow in the occlusion area with the rigid flow obtained by the depth and pose significantly improves the result, which makes the optical flow have a better estimation even if the pixels in the first frame cannot be found in the second frame.

In our system, estimation of depth, pose and optical flow can be realized with high quality by joint unsupervised learning from only monocular video sequences without any labeled ground truth. The depth estimation based on a single image is useful for obstacle avoidance and navigation of drones and autonomous robots as a basic perception capability. The pose estimation can realize the position estimation of autonomous robots. The optical flow estimation can provide information for other works like dynamic object tracking and speed estimation of objects. These problems have profound significance in automatic drive and service robotics.

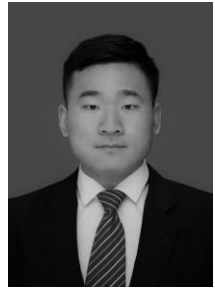

**Guangming Wang** received the B.S. degree from Department of Automation from Central South University, Changsha, China, in 2018. He is currently pursuing the Ph.D. degree in Control Science and Engineering with Shanghai Jiao Tong University. His current research interests include SLAM and computer vision, in particular, depth estimation and pose estimation.

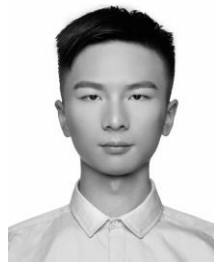

**Chi Zhang** is currently pursuing the B.S. degree in Department of Automation, Shanghai Jiao Tong University. His latest research interests include SLAM and computer vision

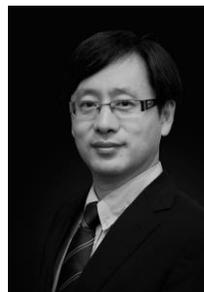

**Hesheng Wang** received the B.Eng. degree in electrical engineering from the Harbin Institute of Technology, Harbin, China, in 2002, the M.Phil. and Ph.D. degrees in automation and computer-aided engineering from the Chinese University of Hong Kong, Hong Kong, in 2004 and 2007, respectively. From 2007 to 2009, he was a Postdoctoral Fellow and a Research Assistant in the Department of Mechanical and Automation Engineering, the Chinese University of Hong Kong. Since 2009, he has been with Shanghai Jiao Tong University, China, where he is currently a Professor in the Department of Automation. His research interests include visual servoing, service robot, robot control, and computer vision. Dr. Wang is an Associate Editor of Robotics and Biomimetics, Assembly Automation, International Journal of Humanoid Robotics, and IEEE TRANSACTIONS ON ROBOTICS. He is the Program Chair of the 2019 IEEE/ASME International Conference on Advanced Intelligent Mechatronics.

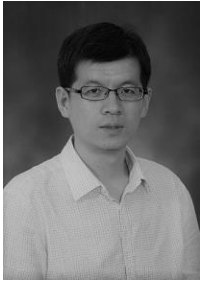

**Jingchuan Wang** received the Ph.D., M.Phil. and B.Eng. degree in Control Theory and Control Engineering from the Shanghai Jiao Tong University, Shanghai, China, in 2002, 2005 and 2014, respectively. Now, he is an Associate Professor in the Department of Automation, SJTU. His research interests include service robot, mobile robot's localization and navigation, and was founded by National Key Research and Development Program of China, Natural Science Foundation of China, Innovation Action Plan of Shanghai and etc.

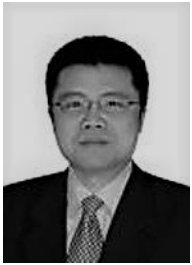

**Yong Wang** received his Ph.D. degree in Control Engineering from Shanghai Jiao Tong University, Shanghai, China, in 2014. Since 2014, he has worked as a senior engineer in Beijing Institute of Control Engineering. His main research interests are space robot, autonomous robot, assistive robot, and collective robot.

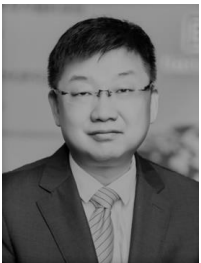

**Xinlei Wang** was awarded doctor degree in Electrical Engineering from Shizuoka University in Japan. He studied under the famous Japanese expert Ikeda Hiroaki in Image and Visual Processing areas. Now, he is currently the Vice President of the Board and the Technical Partner of DeepBlue Technology (Shanghai) Co., Ltd. He is also the Chief Strategy Officer, the Research Director of DeepBlue Science Academy, and the President of the DeepBlue Artificial Intelligence (AI) Chip Research Institute. His research interests include image and video feature extraction, classification and retrieval, pattern recognition, machine vision, deep learning and smart chips, and was founded by Institute of Electrical and Electronics Engineers (IEEE), International Conference on Consumer Electronics (ICCE), etc.